\documentclass[acmtog,nonacm,10pt]{acmart}

\setcopyright{none}
\authorsaddresses{}
\usepackage{amsmath}
\usepackage{booktabs}
\usepackage[ruled]{algorithm2e}
\usepackage{cleveref}
\SetAlFnt{\normalsize}
\SetAlCapFnt{\normalsize}
\SetAlCapNameFnt{\normalsize}
\SetAlCapHSkip{0pt}
\SetAlgoNoLine
\DontPrintSemicolon

\crefname{algocf}{algorithm}{algorithms}
\Crefname{algocf}{Algorithm}{Algorithms}

\begin{document}
\title{Disentangled Stylized Image Generation via IP-Cross-Attention Rearrangement}

\author{Beibei Dong}
\affiliation{%
  \institution{Institute of Automation, Chinese Academy Of Sciences}
  \city{Beijing}
  \country{China}%
}
\email{dongbeibei2022@ia.ac.cn}

\author{Yueming Lyu}
\affiliation{%
  \institution{Institute of Automation, Chinese Academy Of Sciences}
  \city{Beijing}
  \country{China}%
}
\email{lyuyueming2019@ia.ac.cn}

\author{Yue Jiang}
\affiliation{%
  \institution{Institute of Automation, Chinese Academy Of Sciences}
  \city{Beijing}
  \country{China}%
}
\email{jiangyue2021@ia.ac.cn}

\author{Jing Dong}
\affiliation{%
  \institution{Institute of Automation, Chinese Academy Of Sciences}
  \city{Beijing}
  \country{China}%
}
\email{jdong@nlpr.ia.ac.cn}
\renewcommand{\shortauthors}{Dong et al.}

\begin{abstract}
\normalsize
With a solid theoretical foundation and pretraining on large-scale datasets, diffusion-based methods have achieved controllable generation leveraging text and image conditions, and generating excellent stylized images is possible using diffusion models.
However, previous methods face significant challenges.
First, because diffusion generation is iterative, the common approach of learning implicit style representations via cycle-consistency strategies is too costly and not well-suited to diffusion, even for LoRA-based methods.
Second, decoupling style is difficult in the latent space of diffusion models.
Representations of attributes are coupled in the latent space of diffusion models; thus, extracting style attributes with diffusion models usually suffers from the issue of content leakage.
Furthermore, because style is hard to represent in natural language, capturing style using text prompts or textual inversion is deficient.
In this paper, we propose a method for achieving disentangled stylized image generation through IP-cross-attention rearrangement, based on our experimental observations.
We also propose negative content guidance, inspired by classifier-free guidance, to reduce content leakage in stylized image generation.
\end{abstract}

\begin{teaserfigure}
  \centering
  \includegraphics[width=\linewidth]{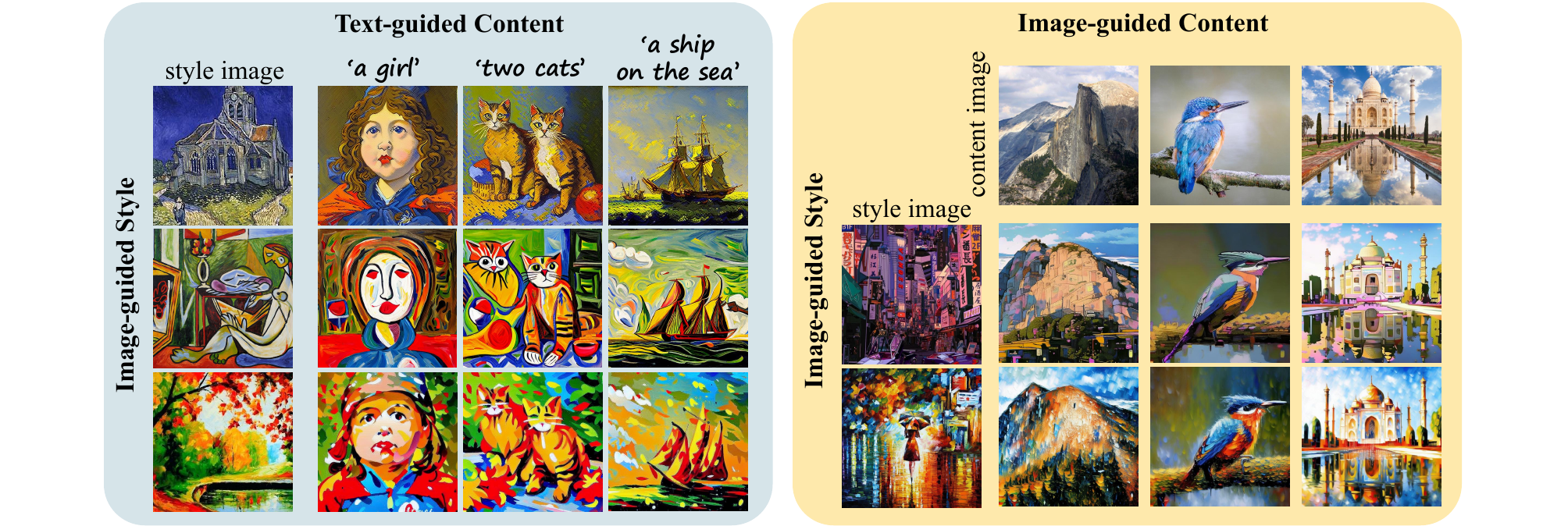}
  \caption{\textbf{Disentangled stylized image generation results using IP-cross-attention rearrangement.} Given different style images, our method can generate results with disentangled style attribute while aligning well with text or image guided content.}
  \label{fig:head}
\end{teaserfigure}

\maketitle
\pagestyle{plain}
\thispagestyle{plain}

\section{Introduction}

Diffusion-based methods have achieved controllable image generation by introducing various forms of conditions, such as text and image conditions.
This controllability enables diffusion models to be applied to multiple tasks like customized and stylized image generation. 

\begin{figure}[h]
  \centering
  \includegraphics[width=0.9\linewidth]{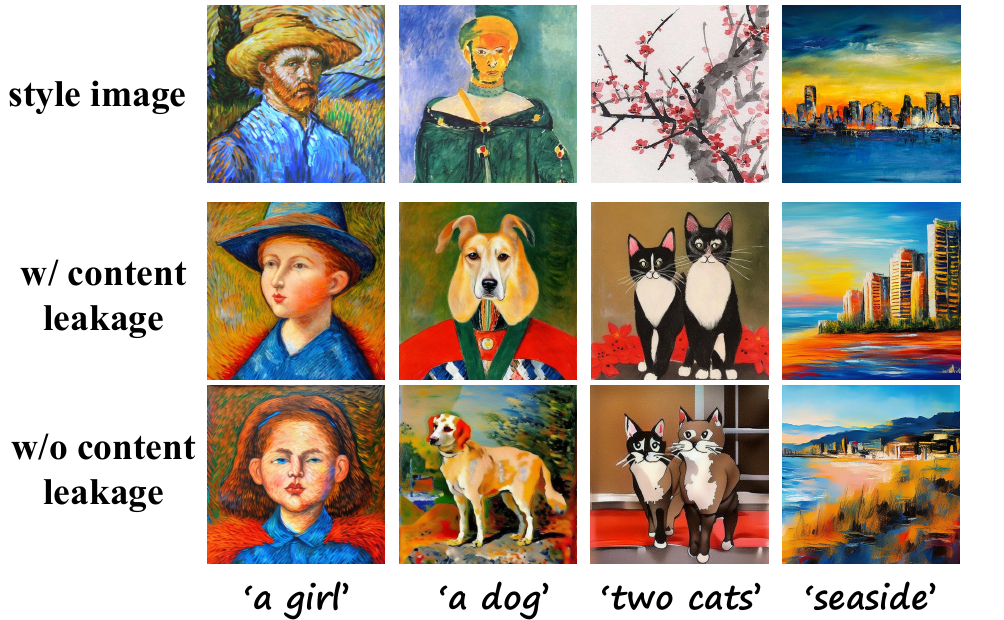}
\caption{\textbf{Content leakage.} We showcase some examples where content leakage occurs. 'w/ content leakage' indicates the presence of content leakage issues. For example, in the generated image of content 'a girl', the content 'hat' in the style image should not appear.}
\label{fig:ip-ada}
\end{figure}

Stylized image generation refers to generating results that match the style of the given style image while aligning with the target content. 
It's crucial to disentangle style and content in stylized image generation.
Previous methods \cite{gatys2016image} \cite{park2019arbitrary} \cite{an2021artflow} \cite{chen2021artistic} \cite{deng2022stytr2} train deep convolutional or transformer-based neural networks to learn the representations of the image content and style. 
Recent investigations leverage the excellent generation capability of the pre-trained diffusion models to achieve stylized image generation with high-resolution. 
Some methods \cite{hertz2023style} \cite{jeong2024visual} \cite{deng2023z} swap or share the key and value features in self-attention blocks to introduce the target style features. 
B-LoRA \cite{frenkel2024implicit} finetune specific blocks within diffusion to separate the style and content implicitly.
To preserve the generative ability of the pre-trained models, some studies \cite{mou2024t2i} \cite{ye2023ip} \cite{wang2023styleadapter} \cite{qi2024deadiff} have explored training additional networks or modules to extract features of the given image.
Although diffusion-based methods can generate high-quality stylized results without requiring training from scratch, they always suffer from the issue of content leakage, which refers to the phenomenon where the content of the style image leaks into the generated result.
The cause of content leakage lies in the entanglement of attributes within the latent space of diffusion models. 
Moreover, diffusion-based methods tend to pay more attention to the image prompts thus the text prompts are ignored. 

To alleviate content leakage, improving the compatibility between the text and the image prompt, we should analyze how diffusion models process these prompts. 
Diffusion models can be seen as a mapper from text to image distributions, as shown in \cref{fig:map} (a).
The introduction of image prompts can be represented by \cref{fig:map} (b), where a mapping from image to image is added.
There are several reasons why diffusion models favor image prompts. 
First, images contain richer information.
When generating images with text prompts, diffusion models only need to query for the most relevant attributes in the learned distribution that match the text. 
Since images contain much more information, the model needs to focus more on the attributes in the image prompt to ensure the accuracy of the query.
Second, it is easier to reconstruct the image prompt. 
Compared to retrieving images based on the text prompt, reconstructing the image prompt makes it easier for the model to receive positive feedback, which leads to the text prompt being overlooked. 
Based on these considerations, a possible solution is to filter the attributes of the image prompt before introducing it into the UNet, as represented by \cref{fig:map} (c), ensuring that only the necessary information of the image prompt is introduced and avoiding masking the information in the text prompt. 

It has been found that text conditions in different cross-attention blocks and timesteps have different impacts on the generated results \cite{zhang2023prospect} \cite{tumanyan2023plug}. Thus, a similar situation might occur with image prompts. 
In this paper, we leverage the pretrained IP-cross-attention module in IP-Adapter \cite{ye2023ip} as the image feature extractor, investigating the impact of image prompts in different blocks and timesteps.
Based on our observations, we propose a block-wise and time-wise IP-cross-attention rearrangement method to achieve disentangled style and content generation. 
Furthermore, leveraging the alignment capability of CLIP \cite{radford2021learning} text and image spaces, we propose a modified classifier-free guidance designed for image conditions using negative content embedding, which can reduce content leakage in generated results.
By disentangling the style and content of the image prompt, our method can not only achieve image-guided stylization but also enables style and content generation guided by different images. 
Our method surpasses existing methods in the disentanglement effect of style and content, ensuring both the style similarity and content alignment. 

\section{Related Work}

\textbf{Diffusion-based text-to-image generation. }
Training on large-scale dataset, diffusion models \cite{ho2020denoising} have achieved excellent generation performance and been used in various downstream tasks. 
Leveraging the CLIP model \cite{radford2021learning}, text-guided image generation methods \cite{kim2021diffusionclip} \cite{ramesh2022hierarchical} \cite{saharia2022photorealistic} \cite{nichol2021glide} \cite{rombach2022high}based on diffusion models can generate images that align well with textual descriptions. 
To meet the higher demands of tasks such as image editing and costomized image generation, researchers have explored to further enhance the controllability of diffusion models. 
Some methods \cite{meng2021sdedit} \cite{kawar2023imagic} \cite{ruiz2023dreambooth} finetune the models to achieve specific tasks like image-to-image translation or image editing. 
To maintain the original generation capabilities of diffusion models, recent studies \cite{zhang2023adding} \cite{mou2024t2i} \cite{ye2023ip} have investigated training additional hypernetworks that adapt to the pretrained models. 
These hypernetworks introduce conditions without compromising the ability of diffusion models generating diverse results, allowing for broader applications. 
Other efforts \cite{hertz2022prompt} \cite{tumanyan2023plug} \cite{brack2023sega} \cite{zhang2023prospect} have explored the information contained in the features of diffusion models, enabling control over the generated results through manipulations of the diffusion features. 

\textbf{Stylized image generation. }
Stylized image generation refers to generating results that match the style of the reference image while aligning with the target content. 
Previous methods \cite{gatys2016image} \cite{park2019arbitrary} \cite{an2021artflow} \cite{chen2021artistic} \cite{deng2022stytr2} use CNN or transformer-based architecture to learn the representations of the image content and style. 
Diffusion-based methods leverage the excellent generation capability of the pre-trained models to achieve stylized image generation. 
Some methods \cite{zhang2023inversion} \cite{wang2023stylediffusion} \cite{deng2023z} use DDIM inversion \cite{song2020denoising} to encode the given image into the latent code of diffusion models to transfer style and content. 
However, inversion-based methods always suffer from insufficient style preservation or attribute entanglement. 
Recent studies \cite{zhang2023adding} \cite{mou2024t2i} \cite{ye2023ip} focus more on training hypernetworks to introduce target image attributes. 
Leveraging the capability of diffusion models to generate text-aligned results, some stylization methods focus on producing stylized images that match the given text prompt. 
Recent studies \cite{hertz2023style} \cite{jeong2024visual} \cite{frenkel2024implicit} \cite{qi2024deadiff} explore the information contained in different features within diffusion models and manipulate target features to introduce style. 
Some of these works \cite{qi2024deadiff} \cite{frenkel2024implicit} \cite{wang2023styleadapter} also refine style control by training or fine-tuning specific modules. 
However, these modules are often trained on a single or a small number of images, which is time-consuming. 
The method \cite{ye2023ip} training hypernetworks provides a great tool for image feature extraction, which also preserve the generalizability of the pre-trained models, thus we can achieve disentangled stylized image generation by modifying it to a style adapter.
\section{Method}
\label{sec:method}

\subsection{Motivation}

\textbf{Content leakage exists in stylized image generation.}
For stylized image generation, it's crucial to disentangle style and content. 
Existing methods using self-attention feature sharing, textual inversion, or model fine-tuning on specific blocks face the challenges of style loss and content leakage.
IP-Adapter leverages CLIP image encoder and a small projection network as an image feature extractor, inserting additional IP-cross-attention blocks into the UNet of the pre-trained diffusion models to process image features and merge with text features. 
Training on 10M images, IP-Adapter can capture image attributes effectively, enabling customized and stylized image generation, etc \cite{ye2023ip}. 
However, the results of the pre-trained IP-Adapter show significant content leakage, as shown in some results in \cref{fig:ip-ada}. 
Moreover, IP-Adapter tends to pay more attention to the image prompts thus the text prompts are ignored. 

To improve the compatibility between the text prompt and the image prompt, we should analyze how diffusion models process these prompts and why the models focus more on image prompts. 

\begin{figure}[h]
  \centering
  \includegraphics[width=0.9\linewidth]{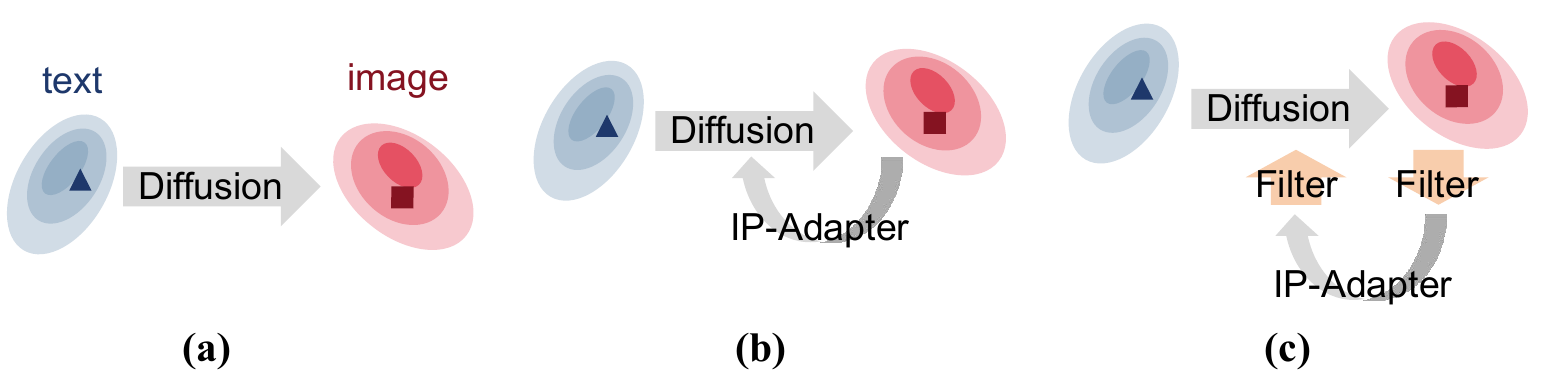}
\caption{\textbf{Modeling of diffusion.} Diffusion models can be seen as a mapper from text to image distribution, while IP-Adapter provides a mapping from image to image distribution, which may cause content leakage.}
\label{fig:map}
\end{figure}

Diffusion models can be seen as a mapper for aligning text and image distributions, as shown in \cref{fig:map} (a).
The introduction of image prompts with IP-Adapter can be represented by \cref{fig:map} (b), where a mapping from image to image is added.
There are several reasons why diffusion models favor image prompts:
(i) Image contains richer information. 
When generating images based on text prompts, diffusion models only need to query for the most relevant attributes in the learned distribution that match the text. 
Since images contain much more information, the model needs to focus more on the attributes in the image prompts to ensure the accuracy of the query.
(ii) Easier alignment of the distributions. 
Due to the similarity between the learned distribution of diffusion models and the distribution of image prompts, compared with text prompts, the models are prone to generate results aligned with the image prompt.

Based on these considerations, a possible solution is to filter the attributes of image prompts before introducing them into the UNet, as represented by \cref{fig:map} (c), ensuring that only the necessary information of the image prompt is introduced and avoiding masking information in the text prompt.

\textbf{Information in text prompts can be selected by controlling the input timesteps and blocks.} 
Zhang et al. \cite{zhang2023prospect} discovered that during the generation process of diffusion, text prompt affects different attributes at different timesteps. 
For example, at smaller timesteps, the text prompt tends to influence attributes such as layout and color of the generated results, while at intermediate timesteps, it controls texture and style more. 
By introducing different text at specific timesteps, attributes of the generated results can be controlled respectively. 
As IP-Adapter merges the extracted image features with text features, disentangled attribute control can be achieved by controlling the input timesteps of image prompts. 

Unlike the original cross-attention maps in UNet, where each channel corresponds to specific text, the IP-cross-attention maps in IP-Adapter consist of 4 channels of image information, and each channel does not correspond to explicit text information. 
Similarly, self-attention maps do not involve any text information, so previous investigation \cite{tumanyan2023plug} focuses more on the differences of self-attention maps in different blocks. 
It has been proved that self-attention maps from different blocks capture different information. 
For IP-cross-attention maps in IP-Adapter, we can analyze the differences between the information contained in different blocks, which might provide a solution to control the generated results. 

\subsection{Observation}
\label{sec:obs}

\begin{figure*}[ht]
  \centering
  \includegraphics[width=\linewidth]{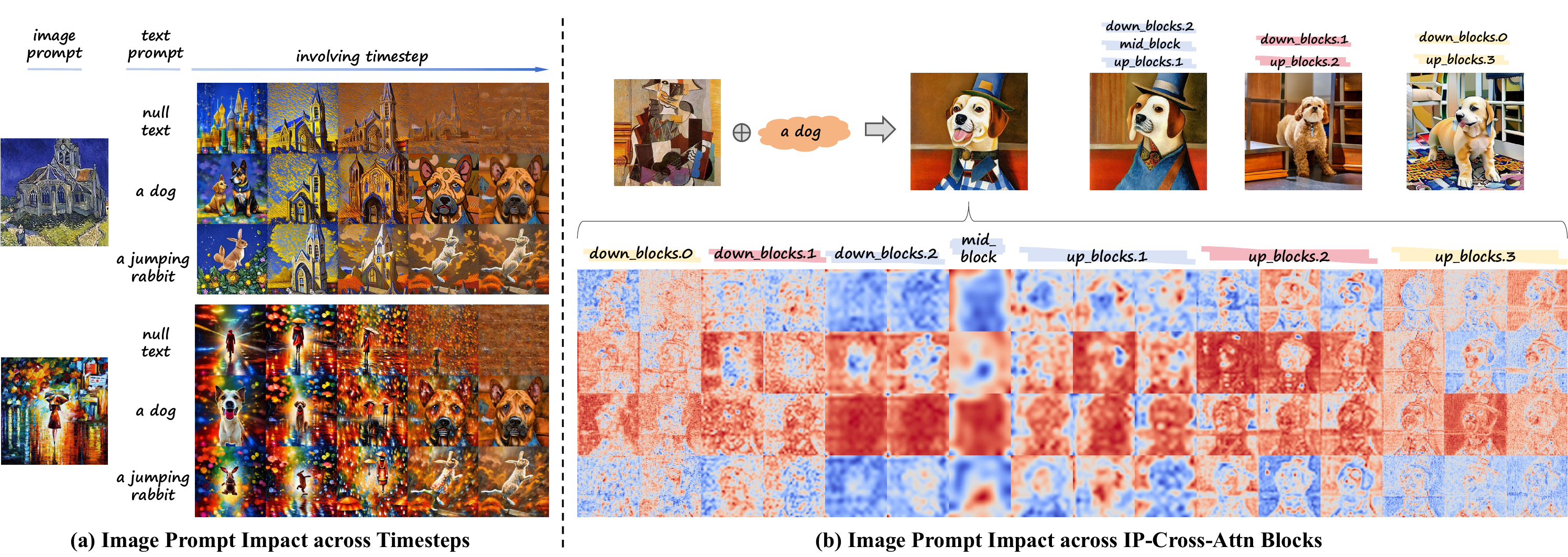}
\caption{\textbf{Results involving image prompts across different timesteps and blocks.} By involving image prompts across different timesteps and blocks, we observed that involving the style image prompt in intermediate and shallow blocks at intermediate timesteps introduces more style information while alleviating content leakage. }
\label{fig:observation}
\end{figure*}

\begin{figure}[h]
  \centering
  \includegraphics[width=\linewidth]{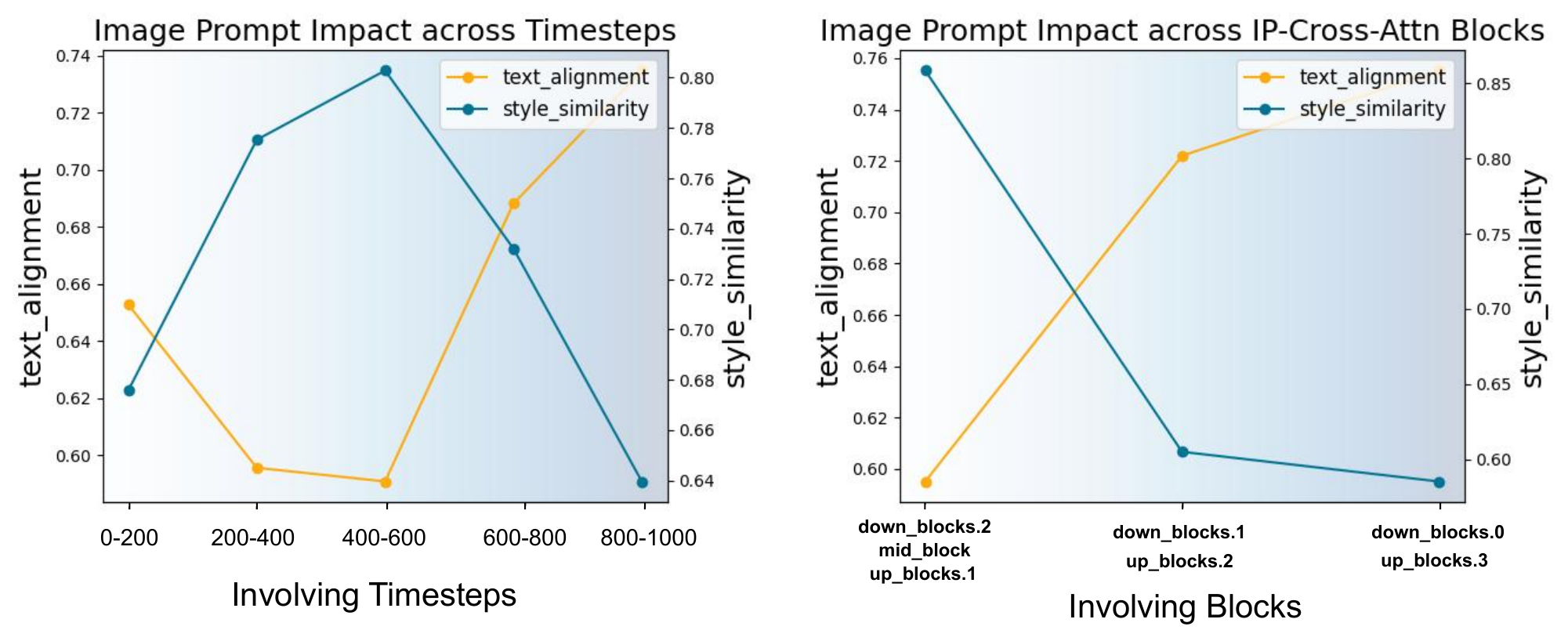}
\caption{\textbf{Analysis with quantitive evaluation. } Involving image prompts at different timesteps and blocks has different impacts, indicated by the trend of text alignment and style similarity metrics.}
\label{fig:zhexian}
\end{figure}

\textbf{Introducing image prompts at different timesteps affects different attributes. }
We conducted experiments introducing image prompts at specific timestep periods. 
As shown in \cref{fig:observation} (a), at smaller timesteps, like 0-300 steps within 1000 steps, introducing the image prompt primarily affects the layout and color of the generated result. 
The introduction of the image prompt at intermediate timesteps mainly involves content and style information, but the layout becomes different from the image prompt. 
At the final 300 steps, involving the image prompt tends to introduce more texture and fine details. 

We further conducted quantitative evaluations to validate our perspective. 
Using pairs of image prompts as the style and text prompts as the content, we introduced the image prompts at different timestep periods, and evaluated the text alignment and style similarity of the generated results using CLIP cosine similarity \cite{radford2021learning} and DINO cosine similarity \cite{caron2021emerging} respectively. 
The results are shown in \cref{fig:zhexian}, which indicates that style is mainly involved at intermediate timesteps, while the content conveyed by text prompts is introduced at earlier and later timesteps.
Thus we can select image prompt attributes by controlling the intensity of the image prompt input at different timesteps.

\textbf{The distribution of attention varies in different IP-cross-attention blocks. }
In \cref{fig:observation} (b), by visualizing the 4 channels of IP-cross-attention maps in IP-Adapter, we can observe that the image prompt has different impacts across different depths of IP-cross-attention blocks. 
As the red areas in attention maps represent higher attention weights, the distribution of the red area reveals the varying levels of attention across blocks. 
For deep blocks like $\textit{down\_blocks.2}$, $\textit{mid\_block}$, and $\textit{up\_blocks.1}$, the distribution of the blue and red areas are more clustered across 4 channels, indicating that the cross-attention maps in deep blocks contain more semantic information. 
Involving the image prompt in these blocks incorporates the layout of image prompts. 
On the other hand, introducing the image prompt in shallow blocks has a more scattered impact, corresponding to more high-frequency information like the texture and details of the generated results.
Similar conclusions can be drawn from the generated results when introducing image prompts in specified blocks.
The quantitative results are shown in \cref{fig:zhexian}.
For deep blocks, the results show poor text alignment, this is because the image prompt has a strong impact on these blocks, making the content of the text ineffective. 
In contrast, for shallow blocks, the generated images align well with the text prompt, but the style similarity drops sharply in intermediate and shallow blocks. 
Based on our observations, we can ensure the style similarity while avoiding the leakage of the style image content by reducing the intensity of the image prompt in the deep blocks and increase it in the intermediate and shallow blocks.

\begin{figure*}[ht]
  \centering
  \includegraphics[width=\linewidth,height=0.82\textheight,keepaspectratio]{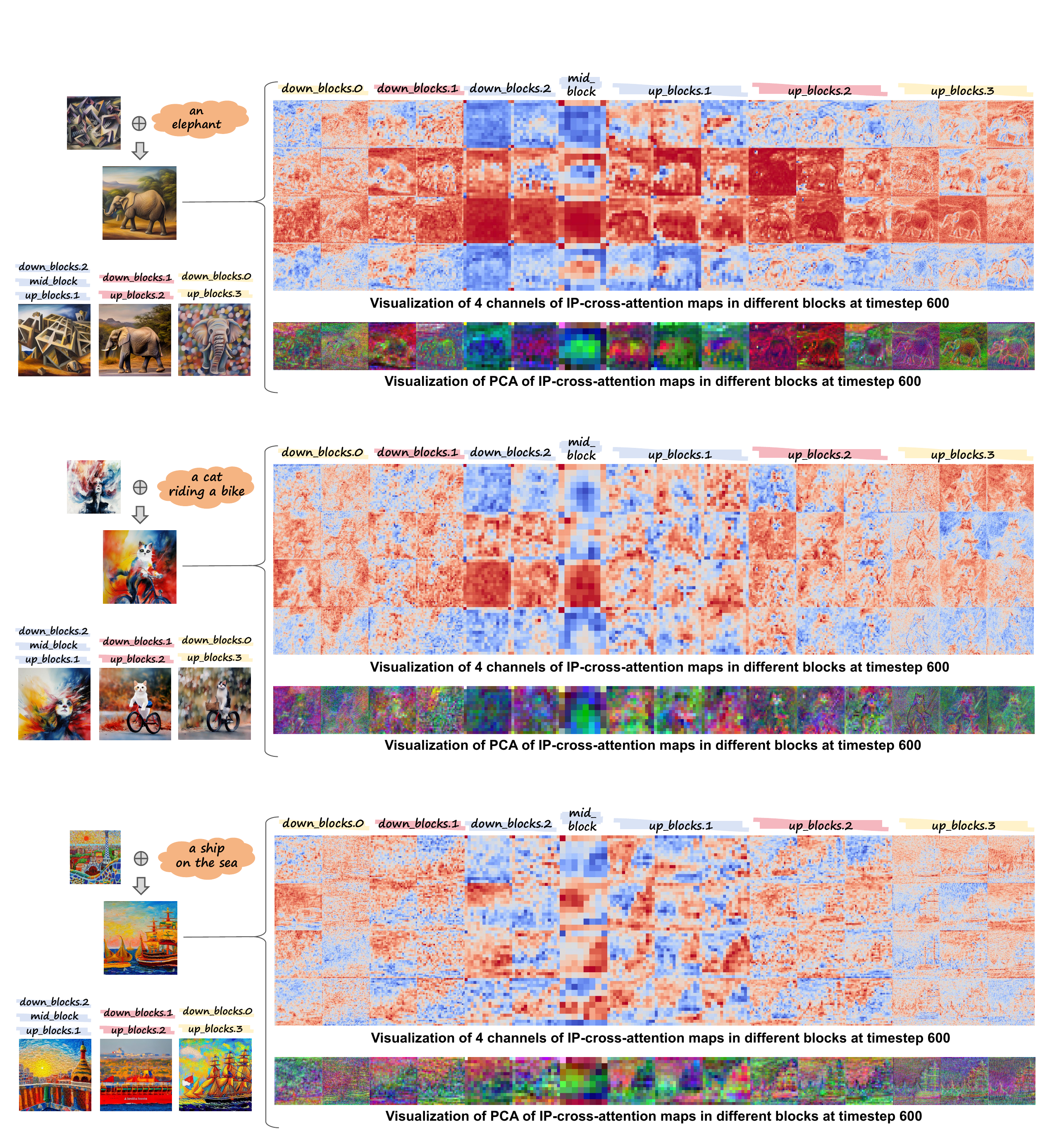}
  \caption{\textbf{Impacts of introducing the image prompt in different blocks.} By visualizing the 4 channels in IP-cross-attention maps in multiple images, we find that deep blocks mainly involve layout and semantic information, while shallower blocks involves more high-frequancy information. We can draw similar conclusion observing the visualization results of PCA analysis.}
  \label{fig:pca}
\end{figure*}

We further applied PCA on the IP-cross-attention maps to visualize the difference of the information contained between different depths of blocks, where we drew a similar conclusion. 
More results are shown in \cref{fig:pca}. 

\subsection{Methodology}
\label{sec:methodology}

\textbf{Image-guided image generation with IP-Adapter.} 
IP-Adapter \cite{ye2023ip} trains image feature extraction modules, including the pre-trained CLIP image encoder, a small projection network, and cross-attention blocks, to introduce image conditions. 
The output of IP-Adapter can be formulated by:
\begin{equation}
\begin{aligned}
\boldsymbol{\epsilon} &= \boldsymbol{\epsilon}_{\theta}(\boldsymbol{x}_t,\varnothing_t,\varnothing_i,t) + {}\\
&\quad s \left(\boldsymbol{\epsilon}_{\theta}(\boldsymbol{x}_t,c_t,c_i,t)-\boldsymbol{\epsilon}_{\theta}(\boldsymbol{x}_t,\varnothing_t,\varnothing_i,t)\right)
\end{aligned}
\label{eq:classi_free}
\end{equation}
$\boldsymbol{\epsilon}$ is the estimated noise at timestep $t$. 
$c_t$ and $c_i$ represent the text and image conditions respectively.
$s$ is the scaling factor of the classifier-free guidance. 
$\varnothing_t$ and $\varnothing_i$ represent the neutral text and image prompts respectively. 
The image features extracted by the IP-Adapter undergo scaling right before being introduced into the UNet: 
\begin{equation}
\begin{aligned}
\operatorname{IP-Cross-Attn}=\operatorname{Attn}\left(Q,K_t,V_t\right) + \lambda \operatorname{Attn}\left(Q,K_i,V_i\right) \\ 
\text{where } \operatorname{Attn}=\operatorname{Softmax}\left(\frac{Q K^{\intercal}}{\sqrt{d}}\right) \cdot V
\label{eq:attn}
\end{aligned}
\end{equation}
The matrices $Q$, $K$, $V$ are obtained by applying the corresponding learned projections to the input features. 
$K_t$, $V_t$ are obtained from text conditions, while $K_i$, $V_i$ are obtained from image conditions.
$\lambda$ is the scaling factor.
IP-Adapter replaces the original cross-attention blocks in the UNet with the IP-cross-attention blocks to achieve image-guided generation.
Although text prompts are compatible with IP-Adapter, the attributes in the image prompt are entangled and the information in text prompts might be covered by the information in image prompts, leading to a content leakage problem. 

\textbf{Disentangle style by rearranging IP-cross-attention.} 
To better utilize the image attribute extraction capability of the pre-trained IP-Adapter and disentangle the target attributes, we can add a filter after the IP-Adapter, as shown in \cref{fig:map} (c). 
The scaling factor $\lambda$ in \cref{eq:attn} can be rearranged into this attribute filter.

As mentioned in \cref{sec:obs}, introduced by IP-Adapter, the image prompt has different impacts at different timesteps. 
Therefore we can filter the target attributes by adjusting the weights of image prompts across timesteps, the rearranged time-wise scaling vector can be represented by:
\begin{equation}
\begin{aligned}
\boldsymbol{\lambda}_t  = \begin{bmatrix}
 \lambda_{1} & \dots & \lambda_{t} & \dots & \lambda_{T}
\end{bmatrix}^\intercal
\end{aligned}
\end{equation}
$\lambda_{t}$ is the scaling factor at timestep $t$, and $T$ is the total timestep which is 1000. 
Similarly, we observed that different IP-cross-attention blocks extract different attributes. 
We can define the rearranged block-wise scaling vector as follows: 
\begin{equation}
\begin{aligned}
\boldsymbol{\lambda}_l  = \begin{bmatrix}
 \lambda_{1} & \dots & \lambda_{l} & \dots & \lambda_{L}
\end{bmatrix}^\intercal
\end{aligned}
\end{equation}
$\lambda_{l}$ is the scaling factor at IP-cross-attention block of index $l$, and $L$ is the amount of IP-cross-attention block in IP-Adapter. 
Combining two rearranged scaling vectors across timesteps and blocks, the final scale $\boldsymbol{\Lambda}$ can be represented in matrix form as:
\begin{equation}
\begin{aligned}
 \boldsymbol{\Lambda} = \boldsymbol{\lambda}_t \cdot \boldsymbol{\lambda}_l^\intercal  = \begin{bmatrix}
 \lambda_{11} & \lambda_{12} & \dots & \lambda_{1L}\\
 \lambda_{21} & \lambda_{22} & \dots & \lambda_{2L}\\
 \vdots & \vdots & & \vdots\\
 \lambda_{T1} & \lambda_{T2} & \dots & \lambda_{TL}
\end{bmatrix}
\end{aligned}
\end{equation}
where $\lambda_{tl}$ represents the scaling factor at timestep $t$ in IP-cross-attention block $l$. 
Replacing the scaling factor in \cref{eq:attn} with our rearranged scale $\boldsymbol{\Lambda}$ enables the control of the image prompt attributes input, thereby achieving attribute-aware generation with IP-Adapter. 

\begin{figure}[ht]
  \centering
  \includegraphics[width=\linewidth]{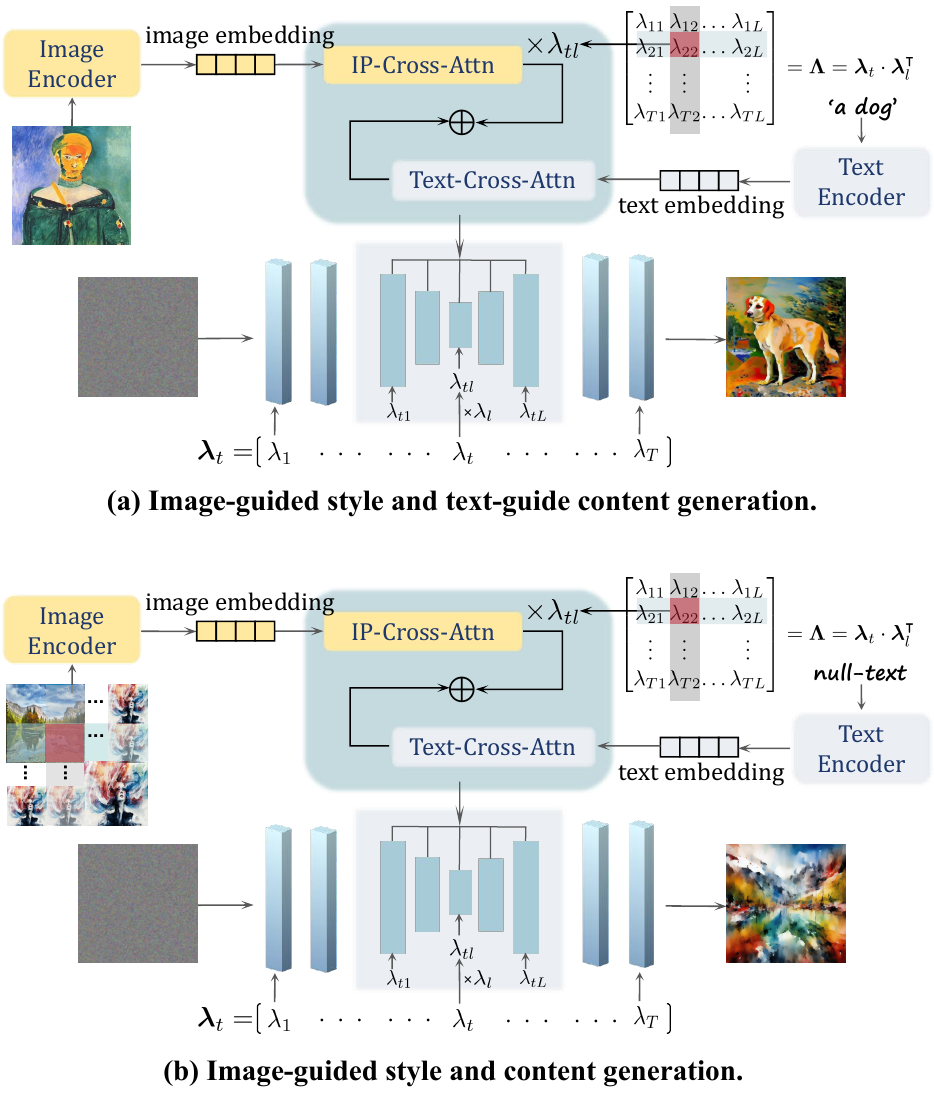}
\caption{\textbf{Our disentangled stylized image generation pipelines.} By introducing the image prompt at timesteps that mainly extract style information with a higher scaling factor, while using a smaller factor in the blocks that mainly capture content information, our method can generate disentangled stylized images with text-guided content.
Moreover, guiding the style and content with different images can also be achieved by introducing specific image prompts in different blocks at specific timestep stages. }
\label{fig:pipeline}
\end{figure}

Considering that style is a combination of finer attributes like texture and color, based on our observations in \cref{sec:obs}, dividing 3-5 stages across timesteps and blocks is sufficient to disentangle style and content.
As shown in \cref{fig:pipeline}, disentangled stylized generation can be achieved by our rearrangement design of the IP-cross-attention scale in IP-Adapter. 
More details about our method including the stage division of timesteps and blocks can be found in implementation details and algorithm in \cref{sec:res}.  

\textbf{Classifier-free guidance with negative content embedding.} 
CLIP model \cite{radford2021learning} is trained using paired text-image data to obtain a sharing latent space aligning text and image representations.
Using text conditions encoded by the CLIP text encoder to control image generation is a common method in diffusion-based image generation. 
As style is hard to convey by natural language, capturing style information using prompts is deficient. 
However, describing image content using natural language is acceptable, and text conditions are usually used to convey content information. 
IP-Adapter, on the other hand, uses the CLIP image encoder as an image feature extractor to introduce image conditions, and the extracted image features contain both the style and content information of the image.
The compatible nature of the CLIP latent space can be leveraged to separate the content within the extracted image features. 

InstantStyle \cite{wang2024instantstyle} uses the text content embedding as negative content embedding and directly subtracts the scaled content embedding from the style embedding in IP-Adapter, and the results show the disentanglement effect of the content. 

To avoid possible deviation of CLIP latent space caused by direct subtraction, and leverage the compatible nature of the CLIP latent space, we propose a modified classifier-free guidance that introduces negative content embedding. 
We use natural language to describe the content of the style image and leverage CLIP text encoder to extract negative content embedding.
We use a parameter $\alpha\in[0,1]$ to balance between neutral and negative prompts:
\begin{equation}
\tilde{\boldsymbol{\epsilon}}=\alpha \boldsymbol{\epsilon}_{\theta}(\boldsymbol{x}_t,\varnothing_t,\varnothing_i,t) + \left(1-\alpha\right)\boldsymbol{\epsilon}_{\theta}(\boldsymbol{x}_t,\varnothing_t,c_n,t)
\label{eq:neg_content}
\end{equation}
$c_n$ is the negative content embedding, which is introduced into the UNet in the same way as the image prompt embedding. 
By replacing $\boldsymbol{\epsilon}_{\theta}(\boldsymbol{x}_t,\varnothing_t,\varnothing_i,t)$ in \cref{eq:classi_free} with $\tilde{\boldsymbol{\epsilon}}$, our modified classifier-free guidance can achieve the disentanglement of style and content in the image prompt. 
\begin{figure}[ht]
  \centering
  \includegraphics[width=\linewidth]{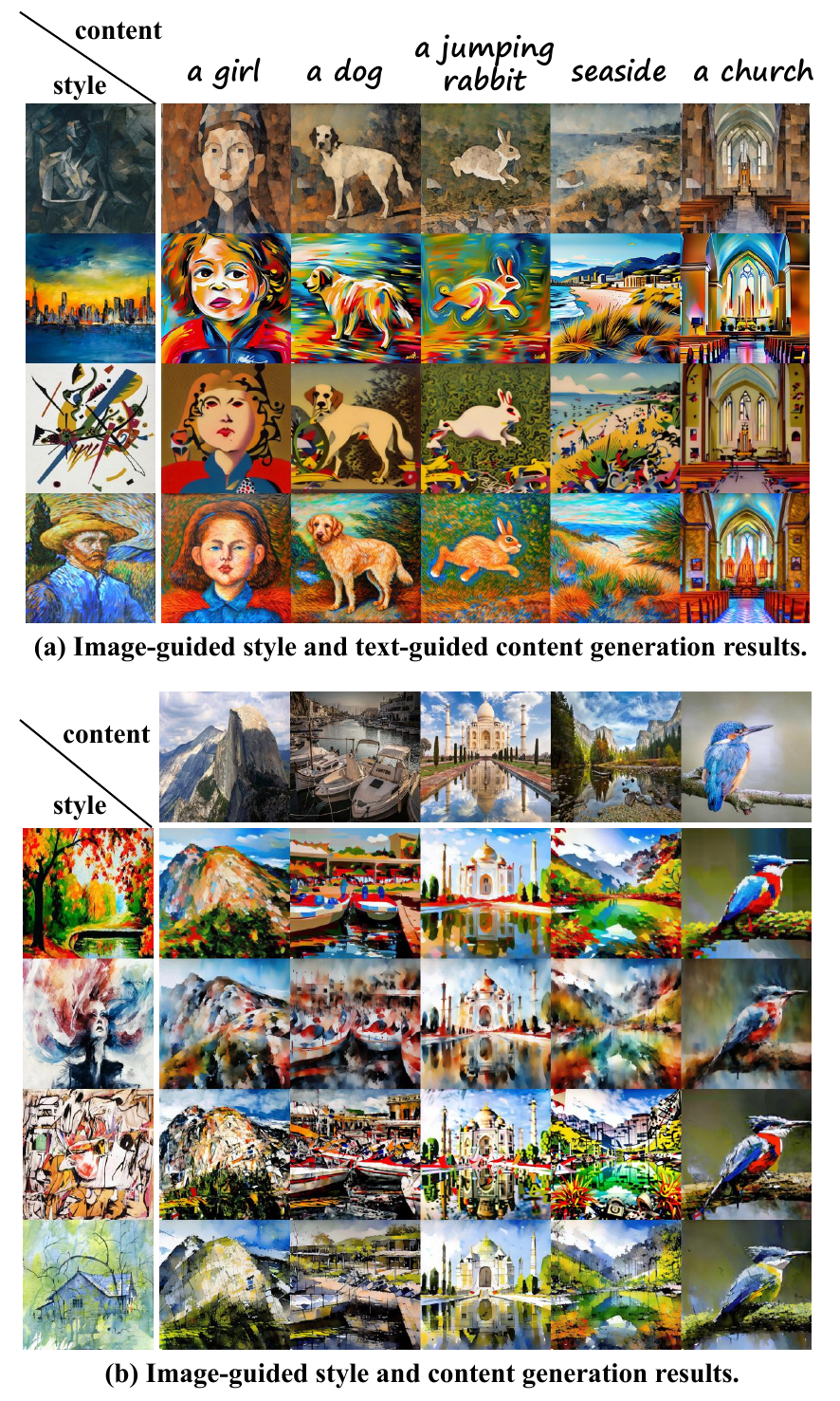}
\caption{\textbf{Some generated results of our method.} Our method can generate disentangled stylized images with text or image-guided content. Image-guided content generation is achieved by introducing content image prompts in the blocks and timesteps that mainly involve content information. }
\label{fig:results}
\end{figure}

\section{Result}
\label{sec:res}

We apply our method described in \cref{sec:methodology} to stylized image generation. 
\cref{fig:results} (a) exhibits the results generated with text-guided content and image-guided style, which is the basic setting of stylized image generation task.
To further demonstrate our observation that early timesteps and deep blocks mainly extract content and layout from the image prompt, we conduct experiments introducing different image prompts across timesteps and blocks.
The results show disentangled style and content guided by the corresponding image prompt, as shown in \cref{fig:results} (b).

It can be found in \cref{fig:results} (b) that the layout of the content image is not precisely transferred into the results.
This is because IP-Adapter only extracts 4 channels of cross-attention features from the image prompt and merges them with the 77 channels of text cross-attention features, thus the information within the image features is insufficient to transfer the layout precisely. 
The general method to accomplish style transfer task, using ControlNet \cite{zhang2023adding} or T2I-Adapter \cite{mou2024t2i} to introduce content from hed or canny sketch of the content image, is also applicable to our method.

Comprehensive comparisons with the state-of-the-art stylized image generation methods are shown in \cref{sec:comp}. 
We also conduct an ablation study to demonstrate the effectiveness of each of our methods.

\begin{figure*}[ht]
  \centering
  \includegraphics[width=\linewidth]{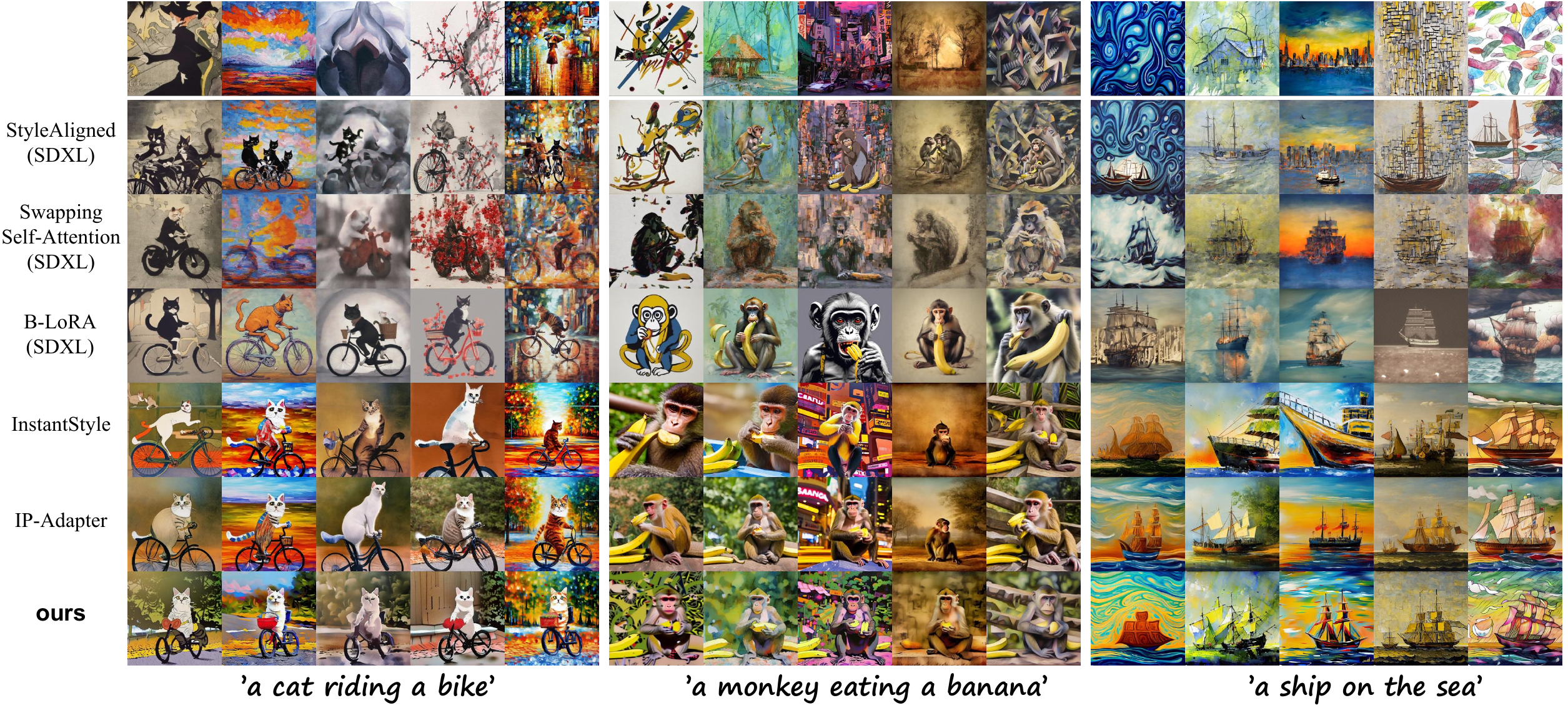}
\caption{\textbf{Comparison with the results of the state-of-the-art methods.} Our method surpasses previous methods with better disentanglement effect of style and content, and the results show excellent style similarity with the style images and align well with the text contents.}
\label{fig:comp}
\end{figure*}

\textbf{Implementation details. }
We use Stable Diffusion 1.5 with 50 timesteps for all of our results.
For time-wise scale rearrangement, we set the timestep thresholds $\boldsymbol{\tau}=\left[15, 25, 50\right]$ and the time-wise scaling vector $\boldsymbol{\lambda_t}=\left[0.9, 2.0, 5.0\right]^\intercal$. 
For block-wise scale rearrangement, we set the block-wise scaling vector $\boldsymbol{\lambda}_l=\left[0.1, 0.4, 0.7\right]^\intercal$ corresponding to IP-Cross-Attention block sets
\begin{equation*}
\begin{aligned}
\boldsymbol{\Theta} = \big[ &[\texttt{down\_blocks.2},\texttt{mid\_block},\\
&\quad\texttt{up\_blocks.1}],\\
&[\texttt{down\_blocks.1},\texttt{up\_blocks.2}],\\
&[\texttt{down\_blocks.0},\texttt{up\_blocks.3}]\big].
\end{aligned}
\end{equation*}
We use $\alpha=0.7$ in our classifier-free guidance with negative content embedding. 
More details about our method are outlined in \cref{alg}.

\begin{algorithm*}[t]
  \caption{Disentangled Stylized Image Generation via Rearranged IP-Cross-Attn}
  \label{alg}
  \KwIn{$\boldsymbol{x}_T$: random noise; $\boldsymbol{\lambda}_t$: time-wise scaling factors;
    $\boldsymbol{\lambda}_l$: block-wise scaling factors; $\boldsymbol{\tau}$: timestep thresholds;
    $\boldsymbol{\Theta}$: IP-Cross-Attention block sets;
    $c_t$: text condition; $c_i$: image condition; $c_n$: negative condition}
  $\boldsymbol{x}_t=\boldsymbol{x}_T$\;
  \For{$t=T \ldots 1$}{
    $\triangleright$ Rearrange scale\;
    \For{$i=1 \ldots \text{length}(\boldsymbol{\tau})$}{
      \If{$t<\boldsymbol{\tau}\left[i\right]$}{
        $\lambda_t=\boldsymbol{\lambda_t}\left[i\right]$\;
      }
    }
    \For{$\operatorname{IP-Cross-Attn} \in \operatorname{UNet}$}{
      \For{$j=1 \ldots \text{length}(\Theta)$}{
        \If{$\text{IP-Cross-Attn} \in \Theta\left[j\right]$}{
          $\text{IP-Cross-Attn}\leftarrow\lambda_{tl}=\lambda_t\times\boldsymbol{\lambda_l}\left[j\right]$\;
        }
      }
    }
    $\triangleright$ Negative content embedding guidance\;
    $\tilde{\boldsymbol{\epsilon}}\leftarrow\alpha \cdot\boldsymbol{\epsilon}_{\theta}(\boldsymbol{x}_t,\varnothing_t,\varnothing_i,t)
      + \left(1-\alpha\right)\boldsymbol{\epsilon}_{\theta}(\boldsymbol{x}_t,\varnothing_t,c_n,t)$\;
    $\boldsymbol{z}_{t-1}\leftarrow\tilde{\boldsymbol{\epsilon}} + s \cdot\left(\boldsymbol{\epsilon}_{\theta}(\boldsymbol{x}_t,c_t,c_i,t)-\tilde{\boldsymbol{\epsilon}}\right)$\;
    $\boldsymbol{x}_{t-1}=\operatorname{DDIMSample}(\boldsymbol{x}_t, \boldsymbol{z}_{t-1}, t)$\;
  }
  \BlankLine
  $\triangleright$ IP-Cross-Attn block merges text and image conditions\;
  \textbf{function} $\operatorname{IP-Cross-Attn}(Q, K_t, V_t, K_i, V_i, \lambda)$\;
  \quad\textbf{return} $\operatorname{Attn}\left(Q,K_t,V_t\right) + \lambda \operatorname{Attn}\left(Q,K_i,V_i\right)$\;
  \textbf{end function}\;
\end{algorithm*}

\subsection{Comparison and Evaluation}
\label{sec:comp}

We compare our method with the state-of-the-art methods, including IP-Adapter \cite{ye2023ip}, InstantStyle \cite{wang2024instantstyle}, B-LoRA \cite{frenkel2024implicit}, Swapping self-Attention \cite{jeong2024visual} and StyleAligned \cite{hertz2023style}. 
As a special note, for methods B-LoRA \cite{frenkel2024implicit}, Swapping self-Attention \cite{jeong2024visual} and StyleAligned \cite{hertz2023style}, due to their limited adaptation to the Stable Diffusion XL model, we use Stable Diffusion XL as the base model for these methods. 
For other methods, we used Stable Diffusion 1.5, which is the same as our setting.

In \cref{fig:comp}, we exhibit the stylized generation results of different methods on various style images using three different text contents.
Our method shows good capability of decoupling content and style, whereas previous methods suffer from the content leakage problem.
We have to note that since we use fixed random noise in the experiments, it is reasonable for the same text prompt to have similar layouts under different styles, and this also proves that the layout of the style image is not leaked into the generated results. 
Our method does not compromise the diversity inherent in our baseline, and diverse results can be generated by changing the random seed.

\begin{figure}[h]
  \centering
  \includegraphics[width=\linewidth]{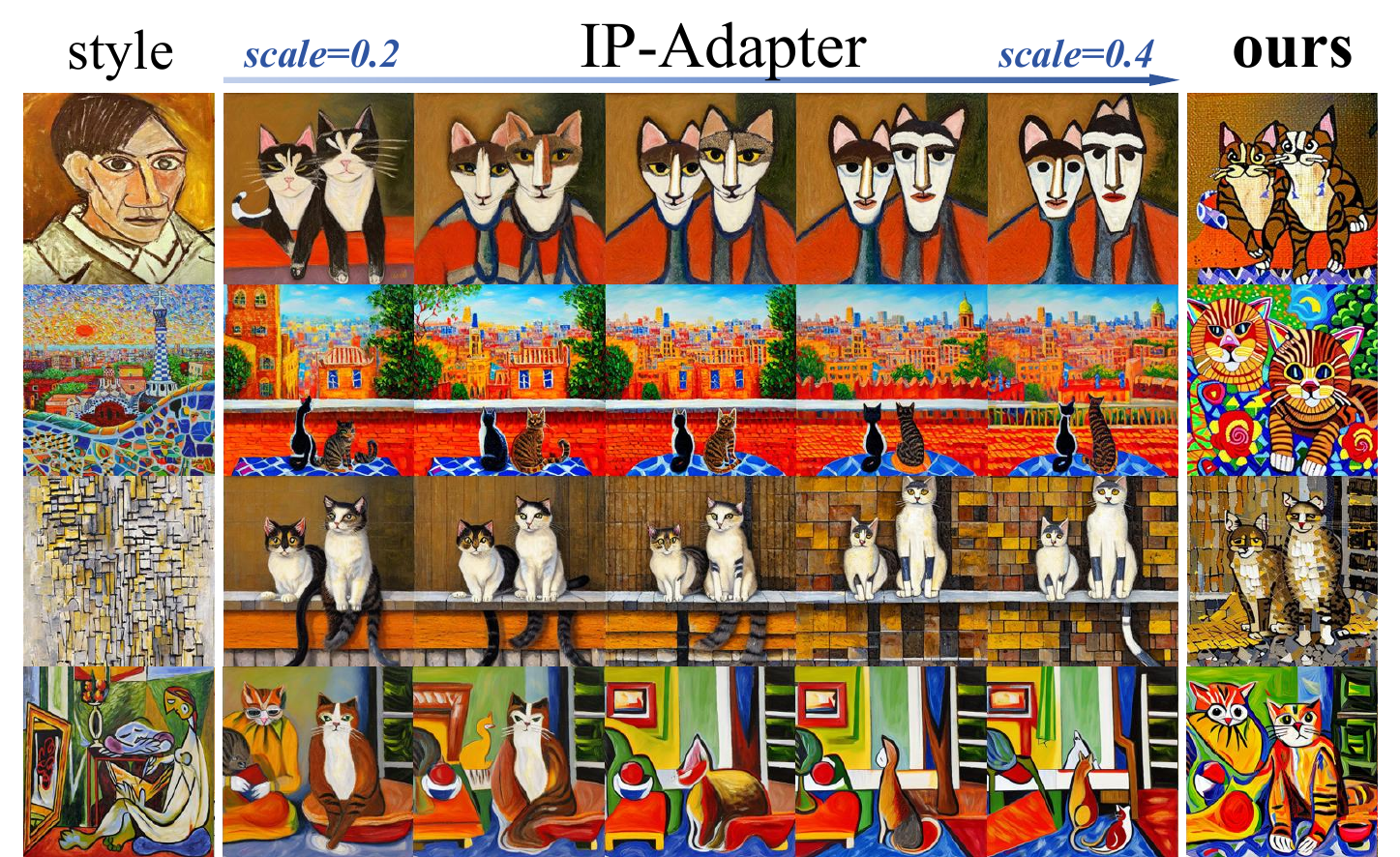}
\caption{\textbf{Comparing with the results of IP-Adapter using different scales.} Under different scales, the results of our baseline IP-Adapter show an unstable presentation of style similarity and text alignment, while our method exhibits good stylization and disentanglement effects on different style images. }
\label{fig:scale}
\end{figure}

Considering that our baseline method IP-Adapter can regulate the global introduction strength of the image prompt by adjusting the scaling factor, we further compare our method with IP-Adapter using different scaling factors to prove that content leakage problem exists in IP-Adapter and demonstrate the effectiveness of our method.
The results are shown in \cref{fig:scale}. 
As the scale of IP-Adapter increases from 0.2 to 0.4, we can observe that the content leakage increases alongside. 
When using smaller scales, some results exhibit insufficient stylization effects, for example, the images in the third row.
In contrast, our method shows relatively stable stylization and disentanglement effects across different style images, verifying the effectiveness of our approach.

\begin{table*}[t]
\centering
\begin{tabular}{@{}cccccc@{}}
\toprule
B-LoRA & SwapAttn & StyleAligned & InstantStyle & IP-Adapter & \textbf{ours} \\
\midrule
16.9\% & 21.2\% & 13.6\% & 10.7\% & 8.4\% & \textbf{29.2\%} \\
\bottomrule
\end{tabular}
\caption{\textbf{User study results.} We conducted a user study to compare with alternative approaches on the disentanglement effect of style and content. Our method attains the highest preference rate, showcasing the alleviation effect of content leakage of our method while ensuring the style similarity and text alignment.}
\label{table:us}
\end{table*}

For quantitative evaluation, considering that there are no suitable metrics for the disentanglement effect of style and image, and the variation in the base model across different methods can result in varying levels of generation quality, we conduct a user study for evaluation.
With several pairs of style images and text contents randomly given, the subjects were asked to select the best 3 results among six results in each pair considering the best disentanglement effects, while ensuring style similarity and text alignment.
We calculated the scores for each method and converted them to a percentage scale. 
As shown in \cref{table:us}, our method achieves the best disentanglement effect of content and style with the highest preference rate.

\subsection{Ablation Study}
\label{sec:abl}

\begin{figure}[h]
  \centering
  \includegraphics[width=\linewidth]{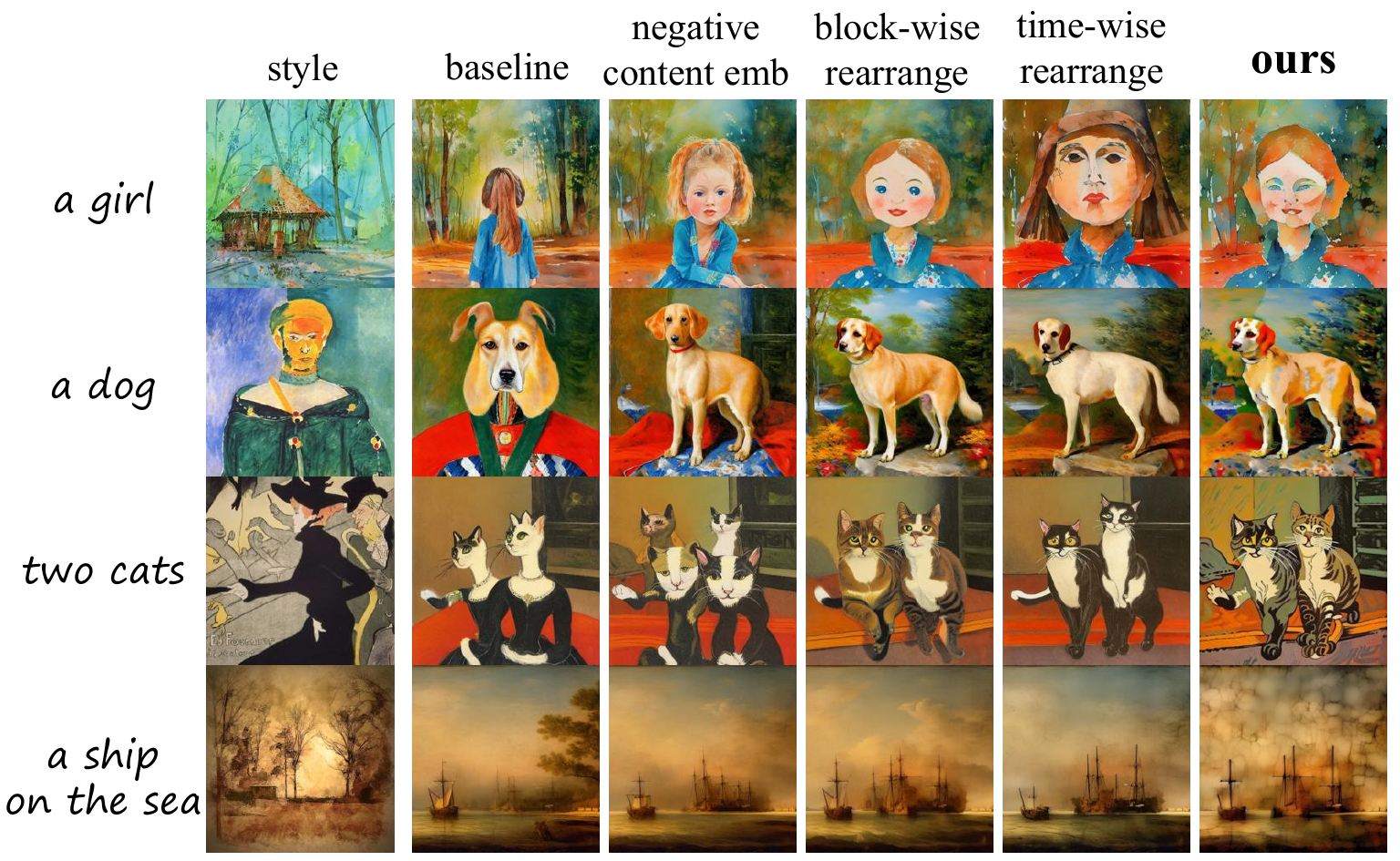}
\caption{\textbf{Ablation study.} With the techniques of our method applied separately, the results show that all techniques can alleviate the content leakage, and our method combining all techniques exhibits the best disentanglement effect of style and content.}
\label{fig:abl}
\end{figure}

Our approach primarily includes three techniques for decoupling style and content, which are (i) Classifier-free guidance with negative content embedding; (ii) block-wise scale rearrangement, and (iii) time-wise scale rearrangement.
To further substantiate our method, we conduct an ablation study where we separately apply the components of our method (i) to (iii), shortened by (i) negative content emb, (ii) block-wise rearrange and (iii) time-wise rearrange, and our method with all components. 
As shown in \cref{fig:abl}, compared to our baseline method, all components of our method show mitigated content leakage, and our method with all components in the last column achieves the best disentanglement effect of style and content.

\section{Conclusion}
In this paper, we propose a method to achieve disentangled stylized image generation via rearranged IP-cross-attention based on IP-Adapter. 

Considering the similarity of the mechanism between image and text conditions in diffusion models, our method is motivated by how previous methods process text conditions to achieve decoupled attribute control.
We conducted experiments to demonstrate that image conditions have different impacts across different timesteps and IP-cross-attention blocks.
Based on our observations, we use time-wise and block-wise scale rearrangement in IP-cross-attention, combined with a modified classifier-free guidance, to generate disentangled stylized images while ensuring style similarity and content alignment. 

Our method surpasses alternative methods on the disentanglement effect of style and content, alleviating the content leakage problem.
Moreover, we provide insights on achieving attribute-aware image generation guided by image conditions, which is a promising research prospect. 

\bibliographystyle{ACM-Reference-Format}
\bibliography{sample-bibliography}

\end{document}